\documentclass[]{spie}  

\usepackage{amsmath,amsfonts,amssymb}
\usepackage{graphicx}
\usepackage[colorlinks=true, allcolors=blue]{hyperref}
\usepackage{booktabs}

\title{Simultaneous Dual-View Mammogram Synthesis Using Denoising Diffusion Probabilistic Models}

\author[a]{Jorge Alberto Garza-Abdala}
\author[a]{Gerardo A. Fumagal-González}
\author[a]{Eduardo de Avila-Armenta}
\author[a]{Sadam Hussain}
\author[b]{Jasiel H. Toscano-Martínez}
\author[a]{Diana S. M. Rosales Gurmendi}
\author[b]{Alma A. Pedro-Pérez}
\author[c]{Jose G. Tamez-Pena}
\affil[a]{ Tecnologico de Monterrey, School of Engineering and Sciences, Monterrey, Mexico}
\affil[b]{ School of Engineering, Pontificia Universidad Católica de Chile, Santiago, Chile}
\affil[c]{ Tecnologico de Monterrey, School of Medicine and Health Sciences, Monterrey, Mexico}

\authorinfo{Further author information: (Send correspondence to Garza-Abdala)\\E-mail: a01376671@tec.mx }

\begin{document} 
\maketitle

\begin{abstract}

\end{abstract}
Breast cancer screening relies heavily on mammography, where the craniocaudal (CC) and mediolateral oblique (MLO) views provide complementary information for diagnosis. However, many datasets lack complete paired views, limiting the development of algorithms that depend on cross-view consistency. To address this gap, we propose a three-channel denoising diffusion probabilistic model (DDPM) capable of simultaneously generating CC and MLO views of a single breast. In this configuration, the two mammographic views are stored in separate channels, while a third channel encodes their absolute difference to guide the model toward learning coherent anatomical relationships between projections. A pretrained DDPM from Hugging Face was fine-tuned on a private screening dataset and used to synthesize dual-view pairs. Evaluation included geometric consistency via automated breast mask segmentation and distributional comparison with real images, along with qualitative inspection of cross-view alignment. The results show that the difference-based encoding helps preserve the global breast structure across views, producing synthetic CC–MLO pairs that resemble real acquisitions. This work demonstrates the feasibility of simultaneous dual-view mammogram synthesis using a difference-guided DDPM, highlighting its potential for dataset augmentation and future cross-view–aware AI applications in breast imaging.
\keywords{Mammography, Breast cancer, Dual-view synthesis, Diffusion models}

\section{INTRODUCTION}
\label{sec:intro}  
Breast cancer remains a major global health  for women, with an estimated 2.3 million new cases and 670,000 deaths reported in 2022 \cite{WHO-Breast-Cancer-2024}. Mammography is currently the gold standard for breast cancer screening and diagnosis due to its proven effectiveness in both early and late stages of detection\cite{Fitzjohn2023}.

Recent advances in machine learning and deep learning have significantly improved the analysis of mammographic images, enhancing the accuracy of breast cancer detection \cite{Khan24}. In particular, dual-view models—those that incorporate both the craniocaudal (CC) and mediolateral oblique (MLO) views—have shown improved diagnostic performance over single-view approaches by leveraging complementary anatomical information \cite{math10234610, wang}. However, these models require large, diverse datasets to achieve robust performance, and many publicly available mammography datasets lack complete dual-view pairs\cite{Sutjiadi_et_al_2024}.

To mitigate this limitation, generative approaches such as generative adversarial networks (GANs) and diffusion models have been employed to synthesize mammographic images. These efforts typically focus on generating single-view images or inferring a missing complementary view \cite{Maistry2023, Yamazaki2022}.

In this work, we introduce a novel method for simultaneous dual-view mammogram synthesis using denoising diffusion probabilistic models (DDPMs)\cite{Ho2020}. Our approach encodes paired views into a single RGB image, where each channel represents distinct structural information: the CC view, the MLO view, and their absolute difference. This encoding allows the model to learn cross-view anatomical consistency during training. To our knowledge, this is the first approach to jointly generate both CC and MLO views in a unified framework, aiming to preserve structural coherence between projections and better reflect the characteristics of real patient data.

\section{METHODS}
\subsection{Data description and preprocessing}
\label{sec:title}
This study utilized a private dataset from the TecSalud health system, comprising 42,454 unique mammogram studies acquired between 2014 and 2019. The dataset includes standard views for each breast—CC and MLO—and was approved by the institutional ethics committee (Protocol No. P000542-MIRAI-MODIFICADO-CEIC-CR002).

To ensure consistency and reduce variability, we included only the first study per patient, restricted the sample to non-implant cases, and focused on patients with BI-RADS-2 findings. This resulted in a total of 22,040 RGB dual-view mammograms from 11,020 patients, with both CC and MLO views available per case.

Image preprocessing involved normalizing pixel intensities to the [0, 1] range. Left breast views (CC and MLO) were horizontally flipped to match the orientation of the right breast, ensuring directional consistency across the dataset. To address inter-image intensity variations, we applied histogram matching as described in previous work \cite{garza2025}. Finally, all images were resized to 256×256 pixels and stored in 16-bit unsigned integer (uint16) format.

\subsection{RGB-based mammograms}
To enable the generation of consistent dual-view mammograms, we encoded paired views into a single RGB image. Specifically, the red channel corresponds to the CC view, the green channel to the MLO view, and the blue channel represents the absolute pixel-wise difference between the CC and MLO images. This encoding strategy allows the model to simultaneously learn view-specific features and their shared anatomical structures.

\begin{figure} [ht]
   \begin{center}
   \begin{tabular}{c} 
   \includegraphics[height=4.5 cm]{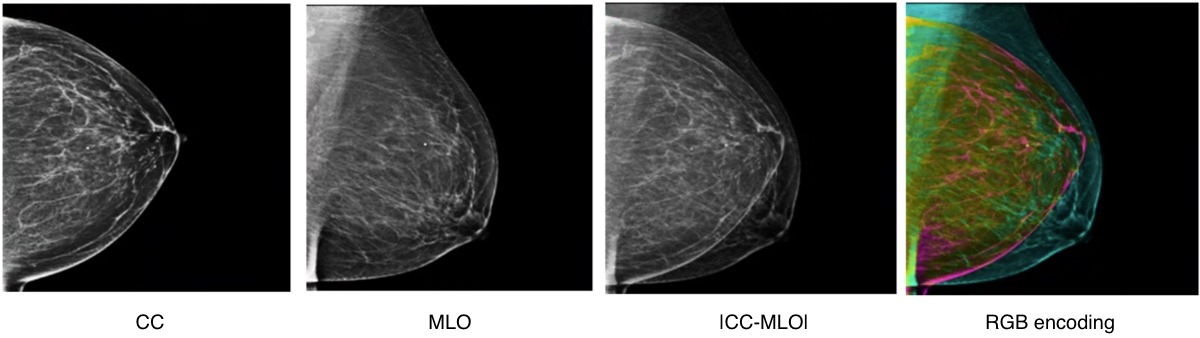}
   \end{tabular}
   \end{center}
   \caption[example] 
   { \label{fig_Real} 
Real mammographic views. From left to right: CC; MLO; $\lvert CC-MLO \rvert$; The three previous views were encoded into an RGB image, where Red=CC, Green=MLO, and Blue=$\lvert CC-MLO \rvert$.}   \end{figure}

As illustrated in Figure \ref{fig_Real}, the resulting RGB image serves as the input to the DDPM, enabling the joint synthesis of both views within a unified framework.

\subsection{Implementation details}
All experiments were conducted using a DDPM on an NVIDIA Tesla A100 GPU. We initialized our model with the publicly available pretrained weights of ddpm-celebahq-256, developed by Google and hosted in the Hugging Face repository \cite{Ho2020}.

To remove residual noise from the generated images, we applied a post-processing step based on percentile normalization. Specifically, the pixel intensities were clipped at the 0.5th and 99.5th percentiles, effectively removing the lowest and highest 0.5\% of values to enhance image quality.

The model was fine-tuned using the following hyperparameters:

\begin{verbatim}
Parameters:
- Learning Rate: 1e-5
- Batch Size: 16
- Epochs: 100
- Scheduler: LambdaLR
- Optimizer:AdamW
- Betas: (0.9, 0.999)
- Loss Function: Mean Squared Error
\end{verbatim}

\subsection{Evaluation}
To evaluate the consistency between synthetic CC and MLO views, we generated 500 synthetic RGB images and conducted a visual inspection test, where a non-expert assessed size, alignment, and artifacts to determine if views appeared from the same patient.

Additionally, we performed a quantitative analysis using binary masks obtained via Otsu’s thresholding, calculating Intersection Over Union (IoU) and Dice Similarity Coefficient (DSC) between views. The same metrics were computed on 2500 real dual-view mammograms from the TecSalud dataset to compare synthetic and real image consistency.

\section{RESULTS}
\label{sec:sections}

\begin{figure} [ht]
   \begin{center}
   \begin{tabular}{c} 
   \includegraphics[height=4cm]{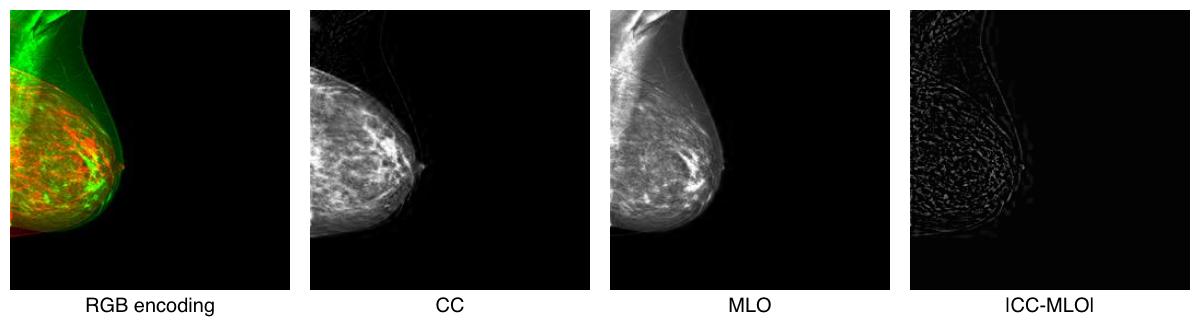}
   \end{tabular}
   \end{center}
   \caption[example] 
   { \label{fig_syn} 
Synthetic mammogram example. From left to right: A synthetic RGB mammogram after being normalized to 99\%; CC view extracted from the synthetic RGB image; MLO view extracted from the synthetic RGB image; $\lvert CC-MLO \rvert$ view extracted from the RGB image.}
   \end{figure}

Figure \ref{fig_syn} presents an example of a synthetic RGB mammogram alongside the individual grayscale views extracted from each channel. The red and green channels correspond to the CC and MLO views, respectively, while the blue channel captures the absolute difference between them. As observed, the model successfully maintains anatomical consistency between the CC and MLO views. The third (difference) view was not used for further analysis in this study.

Following the visual inspection, 94\% of the synthetic images were judged as anatomically consistent, with only minor artifacts attributable to the blending of views. The remaining 6\% exhibited major artifacts, primarily due to preprocessing limitations or inconsistencies in the original dataset.

\begin{figure} [ht]
\begin{center}
\begin{tabular}{c} 
\includegraphics[height=4.5cm]{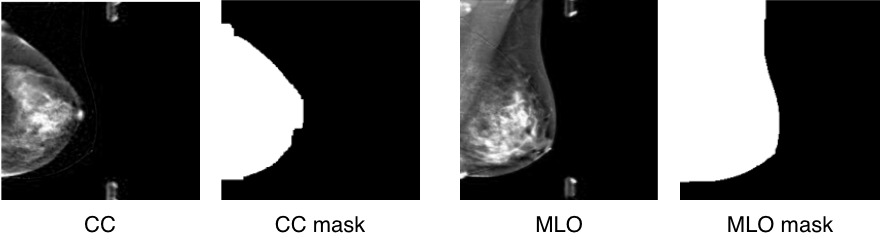}
\end{tabular}
\end{center}
\caption[example] 
{ \label{fig_masks} 
Synthetic mammograms with mask. From left to right: CC view; CC mask; MLO view; MLO mask.}
\end{figure}

\begin{figure} [ht]
\begin{center}
\begin{tabular}{c} 
\includegraphics[width=0.95\textwidth]{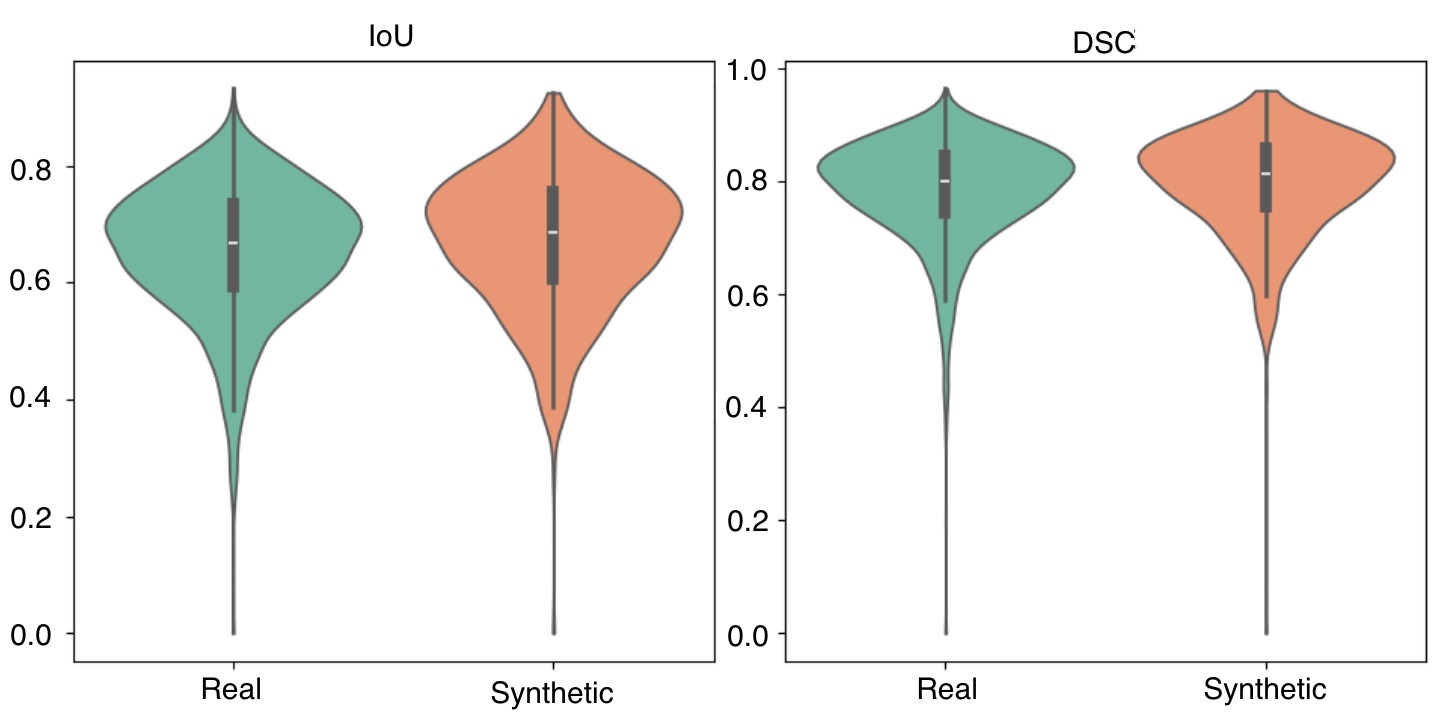}
\end{tabular}
\end{center}
\caption[example] 
{ \label{fig_boxplot} 
IoU and DSC distributions for 2500 pairs of real and 500 pairs of synthetic mammograms' masks.
\\}
\end{figure}

Table \ref{tab:performance_stats_side_by_side} summarizes the descriptive statistics for the IoU and DSC, comparing the binary masks of CC and MLO views. The masks used for this analysis can be seen in Figure \ref{fig_masks}. Figure \ref{fig_boxplot} presents the corresponding violin plots for each metric, illustrating the distribution across both synthetic and real samples.

\begin{table}[ht]
\caption{Performance Statistics for Original and Synthetic Data}
\label{tab:performance_stats_side_by_side}
\begin{center}    
\begin{tabular}{|l|cc|cc|}
\hline
\textbf{Statistic} & \multicolumn{2}{c|}{\textbf{Original}} & \multicolumn{2}{c|}{\textbf{Synthetic}} \\
& \textbf{IoU} & \textbf{DSC}  & \textbf{IoU} & \textbf{DSC} \\
\hline
Mean      & 0.654 & 0.784  & 0.674 & 0.800 \\
Mean difference (vs real) & N/A & N/A & 0.020 & 0.016\\
Std       & 0.113 & 0.092  & 0.107 & 0.081 \\
Min       & 0.000 & 0.000  & 0.261 & 0.414 \\
Q1 (25\%) & 0.592 & 0.743  & 0.605 & 0.754 \\
Median    & 0.668 & 0.801  & 0.682 & 0.811 \\
Q3 (75\%) & 0.733 & 0.846  & 0.751 & 0.858 \\
Max       & 0.932 & 0.964  & 0.908 & 0.925 \\
IQR       & 0.141 & 0.102  & 0.146 & 0.103 \\
\hline
\end{tabular}
\end{center}
\end{table}

Additionally, Table \ref{tab:statistical_tests} reports the results of two statistical tests—Kolmogorov-Smirnov (KS) and Earth Mover’s Distance (EMD)—used to compare the distributions of the real and synthetic datasets. These tests revealed statistically significant differences in IOU and DICE coefficient distributions ($p < 10^{-19}$), suggesting measurable variation between real and synthetic consistency. In contrast, the overlap metric also showed a statistically significant difference, though to a lesser extent ($p < 0.05$).
 
\begin{table}[h!]          
\caption{KS and EMD tests for IoU and DSC, comparing both real and synthetic mammograms (0.005$<$p$<$0.05).}
\label{tab:statistical_tests}
\begin{center}
\begin{tabular}{lccc}
\toprule 
\textbf{Statistic} & \textbf{IoU} & \textbf{DSC}  \\
\midrule 
\textbf{KS D} & $ 0.077^{*} $ & $ 0.077^{*} $ \\
\midrule
\textbf{EMD} & 0.020 & 0.015 \\

\bottomrule 
\end{tabular}
\end{center}
\end{table}

\section{DISCUSSION}
The results demonstrate that simultaneous dual-view synthesis is feasible using a diffusion-based framework and an RGB encoding strategy. Encoding CC and MLO views together with their absolute difference allowed the model to learn cross-view anatomical relationships and produce paired images that were generally coherent in size, orientation, and breast shape. The consistency observed in the visual inspection suggests that joint generation is achievable even when starting from a pretrained natural-image DDPM, requiring only fine-tuning on mammograms.

While prior studies on mammogram synthesis have shown strong performance, they have predominantly focused on generating single-view images or estimating the missing view from an existing one. GAN-based image-to-image models, for example, have demonstrated view-translation capabilities, whereas recent diffusion models have been used to generate high-quality full-field mammograms \cite{Yamazaki2022, Montoya_et_al_2024, JOSEPH2024105456, meng2024diffusionmodelbasedposterior}. However, these approaches do not address the challenge of ensuring anatomical consistency between CC and MLO views of the same breast. In contrast, our approach directly encodes both projections within a unified representation, enabling the model to learn cross-view correspondences implicitly during training.

Despite these encouraging results, the method still presents limitations. First, the absolute-difference channel may amplify small misalignments or intensity variations between CC and MLO views, which could explain some of the artifacts seen in the synthetic samples. Additionally, the evaluation primarily focused on global shape consistency; future analyses should include lesion-level or density-specific assessments to better understand clinical realism. Finally, because this work is based on a single encoding strategy, additional encodings—or conditional mechanisms—may further improve fidelity and reduce artifacts.

Overall, this preliminary study suggests that diffusion models hold promise for generating anatomically consistent dual-view mammograms, opening possibilities for data augmentation in dual-view CAD systems. Future work will explore alternative encodings, conditional generation, and external validation across multiple datasets.

\section{CONCLUSIONS}

Dual-view mammogram generation is essential for advancing computer-aided diagnosis in breast cancer, as clinical interpretation often depends on the complementary information provided by the CC and MLO views. In this study, we introduced a novel RGB-based encoding strategy, where each channel represents a distinct anatomical projection—CC, MLO, and their absolute difference. This design enabled the model to learn a joint representation of both views and synthesize them simultaneously, promoting structural and size consistency.

Although the generated images are not yet suitable for clinical deployment, the proposed method demonstrates promising results in terms of anatomical coherence and visual plausibility. Future work will focus on refining the training pipeline to reduce view-specific artifacts and improve image fidelity. We also plan to explore the effect of the third channel for the DDPM, as well as conditional generation using Latent Diffusion Models or Stable Diffusion frameworks, with conditioning variables such as BI-RADS category to enhance clinical relevance. Additionally, we aim to increase the diversity of the dataset by including more BI-RADS categories and to assess model generalizability through external validation across multiple datasets.

\
\acknowledgments 
This research was supported by the Secretaría de Ciencia, Humanidades, Tecnología e Innovación (Secihti), with cloud computing resources provided through Microsoft’s AI for Good Lab.
\bibliography{report} 

@misc{WHO-Breast-Cancer-2024,
    author = {{World Health Organization}},
    title = {{Breast cancer}},
    howpublished = "WHO, 26 March 2024 \url{https://www.who.int/news-room/fact-sheets/detail/breast-cancer}",
    note = "(Accessed: 29 July 2025)"
}

@incollection{Khan24,
author = {Zeba Khan and Madhavidevi Botlagunta and Gorli L. Aruna Kumari and Pranjali Malviya and Mahendran Botlagunta},
title = {Advancements in Machine Learning and Deep Learning for Breast Cancer Detection: A Systematic Review},
booktitle = {Federated Learning},
publisher = {IntechOpen},
address = {Rijeka},
year = {2024},
editor = {Sultan Ahmad and Meshal Alharbi and Sudan Jha and Aleem Ali and Robertas Damaševičius},
chapter = {2},
doi = {10.5772/intechopen.1008207},
url = {https://doi.org/10.5772/intechopen.1008207}
}

@Article{math10234610,
AUTHOR = {Busaleh, Mariam and Hussain, Muhammad and Aboalsamh, Hatim A. and Fazal-e-Amin and Al Sultan, Sarah A.},
TITLE = {TwoViewDensityNet: Two-View Mammographic Breast Density Classification Based on Deep Convolutional Neural Network},
JOURNAL = {Mathematics},
VOLUME = {10},
YEAR = {2022},
NUMBER = {23},
ARTICLE-NUMBER = {4610},
URL = {https://www.mdpi.com/2227-7390/10/23/4610},
ISSN = {2227-7390},
DOI = {10.3390/math10234610}
}

@inproceedings{wang,
author = {Wang, Zhiwei and Xian, Junlin and Liu, Kangyi and Li, Xin and Li, Qiang and Yang, Xin},
title = {Dual-view correlation hybrid attention network for robust holistic mammogram classification},
year = {2023},
isbn = {978-1-956792-03-4},
url = {https://doi.org/10.24963/ijcai.2023/168},
doi = {10.24963/ijcai.2023/168},
booktitle = {Proceedings of the Thirty-Second International Joint Conference on Artificial Intelligence},
articleno = {168},
numpages = {9},
location = {Macao, P.R.China},
series = {IJCAI '23}
}

@INPROCEEDINGS{Sutjiadi_et_al_2024,
  author={Sutjiadi, Raymond and Sendari, Siti and Herwanto, Heru Wahyu and Kristian, Yosi},
  booktitle={2024 International Conference on Information Technology Systems and Innovation (ICITSI)}, 
  title={Generating High-quality Synthetic Mammogram Images Using Denoising Diffusion Probabilistic Models: a Novel Approach for Augmenting Deep Learning Datasets}, 
  year={2024},
  volume={},
  number={},
  pages={386-392},
  doi={10.1109/ICITSI65188.2024.10929446}}

@article{Maistry2023,
   author = {Brennon Maistry and Absalom E. Ezugwu},
   month = {5},
   title = {Breast Cancer Detection and Diagnosis: A comparative study of state-of-the-arts deep learning architectures},
   url = {https://arxiv.org/pdf/2305.19937},
   year = {2023}
}

@article{Yamazaki2022,
   author = {Asumi Yamazaki and Takayuki Ishida},
   doi = {10.3390/APP122312206},
   issn = {2076-3417},
   issue = {23},
   journal = {Applied Sciences 2022, Vol. 12, Page 12206},
   month = {11},
   pages = {12206},
   publisher = {Multidisciplinary Digital Publishing Institute},
   title = {Two-View Mammogram Synthesis from Single-View Data Using Generative Adversarial Networks},
   volume = {12},
   url = {https://www.mdpi.com/2076-3417/12/23/12206/htm https://www.mdpi.com/2076-3417/12/23/12206},
   year = {2022}
}

@INPROCEEDINGS{garza2025,
  author={Garza-Abdala, Jorge Alberto and Fumagal-González, Gerardo Alejandro and Bosques-Palomo, Beatriz A and Molina, Mario Alexis Monsivais and Avedano, Daly and Cardona-Huerta, Servando and Tamez-Pena, José Gerardo},
  booktitle={2025 IEEE 38th International Symposium on Computer-Based Medical Systems (CBMS)}, 
  title={Ensemble of Radiomics and Convnext for Breast Cancer Diagnosis}, 
  year={2025},
  volume={},
  number={},
  pages={303-306},
  doi={10.1109/CBMS65348.2025.00067}}

@article{Ho2020,
   author = {Jonathan Ho and Ajay Jain and Pieter Abbeel},
   isbn = {2006.11239v2},
   issn = {10495258},
   journal = {Advances in Neural Information Processing Systems},
   month = {6},
   publisher = {Neural information processing systems foundation},
   title = {Denoising Diffusion Probabilistic Models},
   volume = {2020-December},
   url = {https://arxiv.org/pdf/2006.11239},
   year = {2020}
}

@article{Fitzjohn2023,
   author = {Jessica Fitzjohn and Cong Zhou and J. Geoffrey Chase},
   doi = {10.1016/J.IFACOL.2023.10.472},
   isbn = {9781713872344},
   issn = {2405-8963},
   issue = {2},
   journal = {IFAC-PapersOnLine},
   month = {1},
   pages = {5620-5625},
   publisher = {Elsevier},
   title = {Critical Assessment of Mammography Accuracy},
   volume = {56},
   url = {https://www.sciencedirect.com/science/article/pii/S240589632300839X},
   year = {2023}
}

@Article{Montoya_et_al_2024,
AUTHOR = {Montoya-del-Angel, Ricardo and Sam-Millan, Karla and Vilanova, Joan C. and Martí, Robert},
TITLE = {MAM-E: Mammographic Synthetic Image Generation with Diffusion Models},
JOURNAL = {Sensors},
VOLUME = {24},
YEAR = {2024},
NUMBER = {7},
ARTICLE-NUMBER = {2076},
URL = {https://www.mdpi.com/1424-8220/24/7/2076},
PubMedID = {38610288},
ISSN = {1424-8220},
DOI = {10.3390/s24072076}
}

@article{JOSEPH2024105456,
title = {Prior-guided generative adversarial network for mammogram synthesis},
journal = {Biomedical Signal Processing and Control},
volume = {87},
pages = {105456},
year = {2024},
issn = {1746-8094},
doi = {https://doi.org/10.1016/j.bspc.2023.105456},
url = {https://www.sciencedirect.com/science/article/pii/S1746809423008893},
author = {Annie Julie Joseph and Priyansh Dwivedi and Jiffy Joseph and Seenia Francis and Pournami P.N. and Jayaraj P.B. and Ashna V. Shamsu and Praveen Sankaran},
}

@misc{meng2024diffusionmodelbasedposterior,
      title={Diffusion Model Based Posterior Sampling for Noisy Linear Inverse Problems}, 
      author={Xiangming Meng and Yoshiyuki Kabashima},
      year={2024},
      eprint={2211.12343},
      archivePrefix={arXiv},
      primaryClass={cs.LG},
      url={https://arxiv.org/abs/2211.12343}, 
}
\bibliographystyle{spiebib} 

\end{document}